\documentclass[10pt,twocolumn,letterpaper]{article}

\usepackage[pagenumbers]{cvpr} 

\usepackage[dvipsnames]{xcolor}

\newcommand{\medium}{\fontsize{11}{13}\selectfont}

\definecolor{cvprblue}{rgb}{0.21,0.49,0.74}
\usepackage[pagebackref,breaklinks,colorlinks,citecolor=cvprblue]{hyperref}

\usepackage{cuted}
\usepackage{caption}

\def\paperID{394} 
\def\confName{3DV\xspace}
\def\confYear{2026\xspace}

\title{DenseHap: Dense Hand Haptic Estimation from Monocular Images}

\author{%
\begin{tabular}{cc}
Junhan Xue & Rolandos Alexandros Potamias \\
Imperial College London & Imperial College London \\[4pt]
\multicolumn{2}{c}{\href{https://huggingface.co/spaces/Xgxx00200/JointHand}{\medium{\texttt{Project Webpage}}}}
\end{tabular}}
\begin{document}

\vspace{-1.2cm}

\maketitle

\begin{strip}
    \centering
    \vspace{-0.1cm}\includegraphics[width=\linewidth]{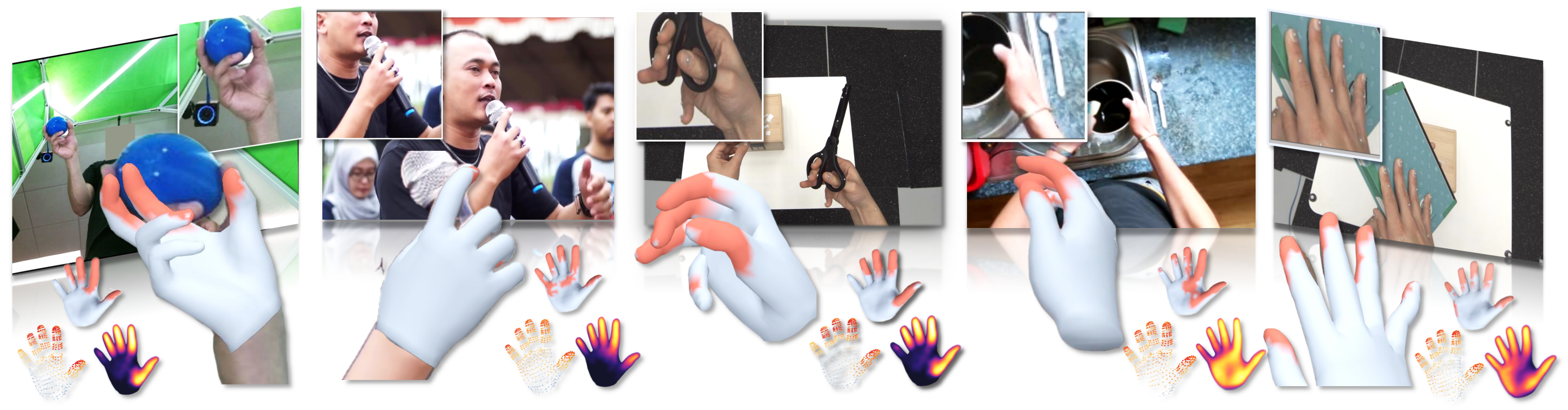}

    \captionsetup{hypcap=false}
    \captionof{figure}{We propose DenseHap for hand distance field estimation, contact detection, pressure analysis, and reconstruction from a single monocular image. DenseHap improves HACO's F1 by about 10\% on zero-shot datasets with only 40M additional trainable parameters.}
    \label{fig:overview}
    \captionsetup{hypcap=true}
\end{strip}

\begin{abstract}

As dexterous robotic manipulation advances, scalable annotations of human hand-object interactions are becoming increasingly important. Furthermore, previous hand contact estimation methods suffered from issues such as dorsal-side misclassifications and insufficient precision, while lacking a unified framework for integrating multiple hand-to-environment interaction modalities, including distance fields, contact thresholds, and pressure. To address that, we present DenseHap, a unified end-to-end approach to estimate hand distance field recognition, comprehensive hand contact detection, hand pressure analysis, and hand reconstruction from a single monocular image. By incorporating three distinctly defined heads, along with a pretrained 3D reconstructed skeleton and an enhanced dataset, DenseHap outperforms previous state-of-the-art models on four evaluation datasets, improving the F1 score by 8.5 percentage points (a 16\% relative improvement) on untrained datasets and by 3 percentage points (5\% relative improvement) on in-distribution datasets. Furthermore, DenseHap has effectively improved short-range hand estimate and pressure evaluation. 
Code and models will be publicly available upon the paper's acceptance.  

\end{abstract}    
\section{Introduction}
\label{sec:intro}

Recent dexterous manipulation systems~\cite{zhao2026dexmachina, pan2025spider,wu2026cedex,wu2026unimorphgrasp} offer increasing proof that contact serves as an efficient medium for transmitting human demonstrations to robotic hands. Simultaneously, the acquisition of dexterous manipulation skills increasingly depends on human videos for guidance. Research such as DexMV~\cite{qin2022dexmv}, VideoDex~\cite{shaw2023videodex}, and DexVIP~\cite{mandikal2022dexvip} illustrate the transformation of hand trajectories from these videos into robotic policies, underscoring the considerable opportunities presented by the abundant online resources of hands interacting with objects.

However, these works typically concentrate on hand motion trajectories, while the primary factor that confers physical significance to manipulation, namely the point at which the hand makes contact with the world, is either disregarded or approximated post-hoc. This missing supervisory signal is precisely provided by dense hand contact. In this work, it is defined as the contact state at each vertex of the MANO mesh \cite{romero2017mano}, including point-wise distance fields, binary contact indicators, and force magnitudes. This signal grounds the reconstructed hand in the physical reality of its interactive behaviour. Therefore, as an explicit form of information, it provides direction for the subsequent disentanglement of hand motion representations and the fine-grained comprehension of manipulation details.

In terms of efficiency, current approaches for dense contact estimation necessitate the use of a cascade of two independent models. Dense contact estimators, such as HACO~\cite{jung2025haco}, predict per-vertex contact probabilities from cropped hand images but recover no geometry. Monocular hand reconstruction methods~\cite{pavlakos2024hamer,potamias2025wilor,chen2025hort,si2026anyhand} recover MANO meshes accurately and at high throughput, but ignore contact. As a result, the estimator never views the mesh that its contact map is intended to annotate. In addition to requiring two inference passes, this separation allows for physically inconsistent predictions. For example, a fingertip placed at a significant distance from a surface that the estimator indicates as touched, without a mechanism to reconcile the two, renders the pipeline both unreliable and costly at the scale of millions of video frames. Moreover, hand contact is more fine-grained than full-body contact, often occurring in small regions such as the fingertips, which limits the effectiveness of conventional full-body contact estimators \cite{jung2025haco,chen2025highresolution}.

Motivated by these findings, we present DenseHap, an integrated framework that generates intricate per-vertex hand interaction data from a single RGB image in one forward pass, without the need for object templates or optimisation during testing. Three heads are connected to a WiLoR-based reconstruction backbone, with each head overseeing a shared visual representation. Furthermore, by employing the ViT backbone of WiLoR, a straightforward head allows for the effortless acquisition of 3D reconstruction results from identical features.

\noindent Our contributions are as follows: 
\begin{itemize}

\item We introduce the first end-to-end framework that forecasts dense per-vertex interaction (distance, contact, pressure and reconstruction) and reconstructs 3D hand geometry from a single RGB image, utilising joint supervision over contact labels, spatial distance, and pressure signals.
\item Our approach attains cutting-edge dense contact precision, functioning at approximately double the throughput of current reconstruction-and-contact cascade systems, showcasing enhanced accuracy and efficiency.
\item 
We examine a collection of prior studies and perform a basic credibility assessment of these datasets on the contact dimension. Simultaneously, using the data sampling from the HACO data set, we developed an enhanced data cleaning procedure, resulting in a dataset that was double the size of the original.

\end{itemize}

\section{Related Work}
\label{sec:Related_Work}

\noindent\textbf{Hand Contact and Pressure Estimation.}
Early work on explicit hand-contact measurement relied on instrumented capture. ContactDB \cite{brahmbhatt2019contactdb} uses thermal imaging to acquire object-surface contact maps after grasping, while ContactPose \cite{brahmbhatt2020contactpose} associates these measurements with hand--object poses and RGB-D observations. At the image level, 100DOH \cite{shan2020100doh} and ContactHands \cite{narasimhaswamy2020contacthands} detect hands and classify their contact states in Internet images, but do not localise contact on the hand surface. Dense 3D contact has also been used as an intermediate representation for physically plausible hand--object reconstruction. ContactOpt \cite{grady2021contactopt} optimises hand pose using predicted contact, while CPF \cite{yang2021cpf}, TOCH \cite{zhou2022toch}, S\textsuperscript{2}Contact \cite{tse2022s2contact}, and UV-based reconstruction \cite{yu2022uvcontact} model dense contact through geometric, temporal, semi-supervised, or UV-space representations. Beyond these reconstruction-oriented methods, POSA \cite{hassan2021posa} learns pose-conditioned contact priors between the body and scene, while BSTRO \cite{huang2022bstro} and DECO \cite{tripathi2023deco} infer dense vertex-level body contact from RGB images; their hand-region predictions provide relevant baselines. ContactPrompt \cite{jung2026contactprompt} further performs training-free dense hand contact estimation through structured MLLM reasoning. Among supervised image-based approaches, HACO \cite{jung2025haco} is the closest general-purpose baseline to our setting. It predicts contact probabilities over the MANO vertices and explicitly addresses class and spatial imbalance. However, HACO treats contact estimation as a standalone task and does not recover input-specific 3D hand geometry, requiring a separate reconstruction network and thereby preventing shared feature learning while increasing the cost of large-scale image or video annotation.
Beyond binary contact, PressureVision \cite{grady2022pressurevision} and PressureVision++ \cite{grady2024pressurevisionpp} estimate pressure from RGB images of interactions with instrumented planar surfaces, while EgoPressure \cite{zhao2025egopressure} extends pressure estimation to egocentric views with hand-pose annotations. More recently, HOPE \cite{jeon2026hope} predicts per-vertex contact and normal pressure from monocular video, and EgoPHI \cite{ilic2026egophi} estimates contact and 3D forces on hand and object meshes, but requires known object geometry. In contrast, our method jointly recovers input-specific hand geometry and dense surface contact from a single RGB image.

\noindent\textbf{Monocular 3D Hand Reconstruction}
 The MANO parametric model \cite{romero2017mano} and increasingly powerful feed-forward architectures have accelerated the recovery of 3D hand geometry from single RGB images. Early systems such as FrankMocap \cite{rong2021frankmocap} demonstrated practical monocular capture of hands and the full body, while transformer-based methods such as HaMeR \cite{pavlakos2024hamer} achieved robust in-the-wild hand mesh recovery through large-scale training and high-capacity backbones. WiLoR \cite{potamias2025wilor} further unifies hand localisation and reconstruction, providing the throughput and accuracy required to process unconstrained video at scale. These properties make modern reconstruction networks attractive for mining Internet video, enabling dynamic world-space reconstructions \cite{zhang2025hawor,zeng2026stablehand} and hand-object interactions \cite{liu2025easyhoi,chen2025hort}. However, they recover only geometry, leaving unmodeled the signals that distinguish manipulation from mere motion and indicate physical contact with the environment.

\begin{figure*}[h]
\begin{center}
 \includegraphics[width=0.97\linewidth]{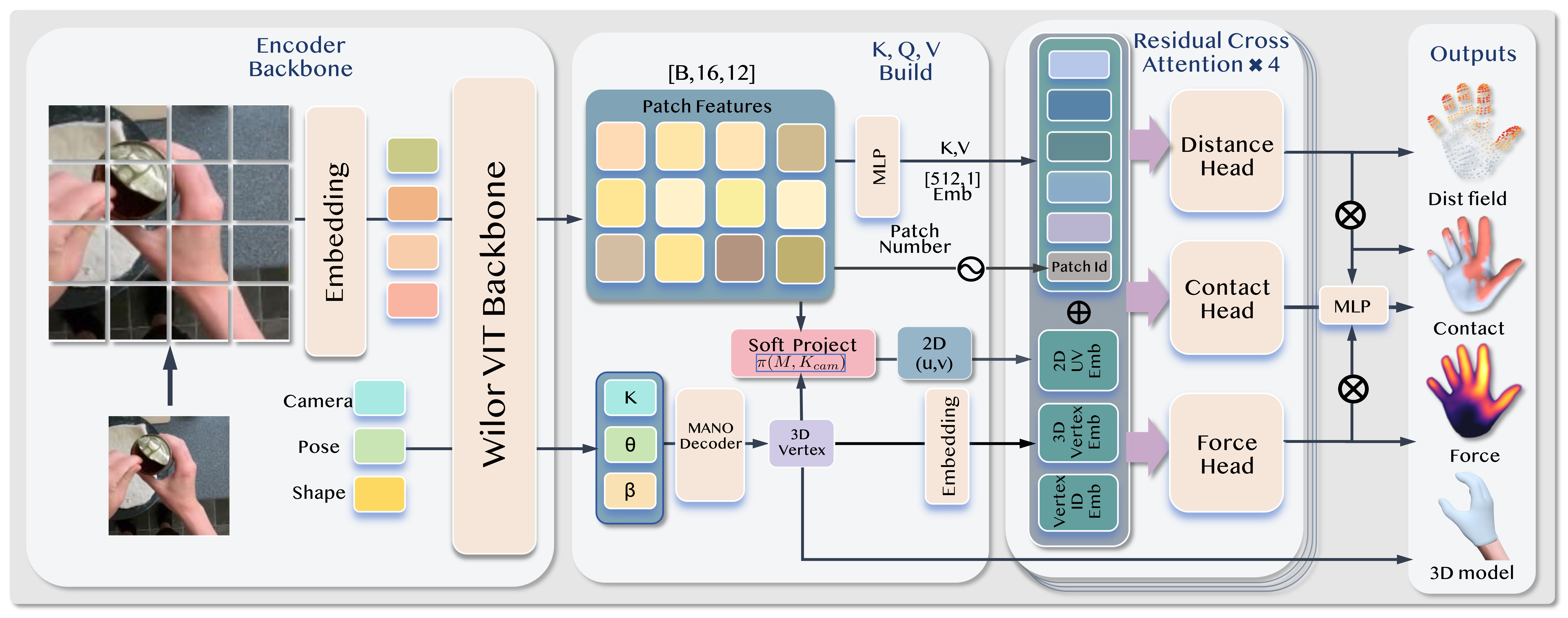}
 \vspace{-5pt} 
\end{center}
   \caption{Overview of DenseHap. A frozen WiLoR \cite{potamias2025wilor} backbone yields the MANO \cite{romero2017mano} mesh
and a ViT feature grid; three parameter-disjoint vertex-query cross-attention
trunks feed contact, distance, and pressure heads under triple supervision.
Detached proximity signals from the distance and pressure heads are injected into
the contact head through zero-initialized learnable gates.}
\label{fig:overall_arch}
\end{figure*}

\section{Methodology}

Given a single RGB image, DenseHap predicts, in one forward pass and without
object templates or test-time optimization, a 3D hand mesh in MANO \cite{romero2017mano} topology
together with a dense per-vertex description of its physical interaction: a
contact probability, a vertex-to-object distance, and a contact pressure at
each vertex. The overall architecture of the proposed model in shown in 
Fig ~\ref{fig:overall_arch}. 

\subsection{Frozen Reconstruction Backbone}

We employ WiLoR as our frozen image encoder and reconstruction framework. The ViT-H/16 architecture \cite{dosovitskiy2020image} simultaneously analyses image tokens along with MANO-associated form, posture, and camera embeddings to yield hand-sensitive patch features, which are then decoded to recreate the 3D hand mesh. Our methodology employs two outputs from WiLoR: the reconstructed MANO vertices and the intermediate ViT feature map. The vertices provide explicit 3D geometric cues for contact reasoning, while the feature map serves as the visual representation for our prediction heads. Given that WiLoR is pre-trained explicitly for 3D hand reconstruction, its attributes are more congruent with hand-contact estimates compared to general semantic representations like DINOv3 \cite{simeoni2025dinov3}.

\subsection{Hybrid Cross-Attention}

The contact head treats the hand vertices \(V\)  as queries that
cross-attend to WiLoR ViT patch tokens. The geometry-aware query serves
as an \emph{addressing prior} that initializes a residual attention
stream rather than constraining each vertex to a fixed image location.
Thus, early layers are guided by anatomical and geometric cues, while
deeper layers progressively refine vertex features using image evidence.

\noindent\textbf{Vertex queries (\(\mathbf{Q}\)).}
Contact at one vertex can depend on distal hand regions, making
independent vertex processing insufficient for capturing global context.
We therefore represent the \(V\) MANO vertices as a set of queries whose
dependencies are modelled by successive attention layers.

Let
\(\mathbf{Q}=[q_1,\ldots,q_V]^\top\in\mathbb{R}^{V\times d}\).
For vertex \(i\),
\begin{equation}
q_i
=
e^{\mathrm{id}}_i
+
\phi_{\mathrm{2D}}\!\left(\pi(v_i)\right)
+
\psi_{\mathrm{xyz}}\!\left(v_i-\bar v\right),
\label{eq:query}
\end{equation}
where $\bar v=\frac{1}{V}\sum_{j=1}^{V}v_j.$

Here, \(e^{\mathrm{id}}_i\in\mathbb{R}^{d}\) is a learnable embedding
encoding vertex identity. The projection \(\pi(v_i)\) uses WiLoR camera
parameters to map the reconstructed vertex onto the normalized ViT patch
grid, and \(\phi_{\mathrm{2D}}\) converts the projected coordinate into a
continuous sinusoidal embedding. The centroid-relative coordinate
\(v_i-\bar v\) is encoded by \(\psi_{\mathrm{xyz}}\) to provide
translation-invariant 3D geometry.

Each query therefore combines anatomical identity, image-aligned 2D
position, and 3D geometric context. These cues initialize where and how a
vertex attends, while subsequent attention layers may refine or override
this initial spatial preference according to image content.

\noindent\textbf{Cross-attention and multi-scale prediction.}
WiLoR ViT patch tokens are projected to width \(d=512\) by an MLP and
augmented with 2D sinusoidal positional encodings. They serve as keys and
values for \(L=4\) pre-norm residual cross-attention blocks:
\begin{equation}
\begin{split}
\tilde{\mathbf Q}^{(l)}
&=
\mathbf Q^{(l-1)}
+
\mathrm{CrossAttn}
\!\left(
\mathrm{LN}(\mathbf Q^{(l-1)}),\mathbf K,\mathbf V
\right), \\
\mathbf Q^{(l)}
&=
\tilde{\mathbf Q}^{(l)}
+
\mathrm{FFN}\!\left(\mathrm{LN}(\tilde{\mathbf Q}^{(l)})\right).
\end{split}
\end{equation}
The residual stream refines the initial
geometry-aware queries while progressively incorporating image-conditioned
evidence, allowing vertices to attend beyond their projected locations.
The refined features are finally used for contact prediction, with
supervision at four resolutions
(\(778,336,84,16\)) for multi-scale consistency.

\subsection{Final Output}
Although we train three heads, the final output can be derived directly from either the distance-threshold head or the contact head. On most datasets, the distance-threshold head performs better. The contact head outperforms it only when training data for the target dataset are unavailable or when the test annotations are human-labelled and widely dispersed. We therefore adopt the distance-threshold head as our default model. Nevertheless, the contact head remains important: it plays a mediating role that benefits semantic segmentation and performs well on datasets with subjective annotations.

\subsection{Loss Function}
\label{sec:triple}

Contact, distance, and pressure provide complementary descriptions of
hand--object interaction. HACO~\cite{jung2025haco} defines contact by thresholding
the vertex-to-object distance:
\begin{equation}
c_{n,v}
=
\mathbb{I}\!\left[
d\!\left(\mathbf{x}^{\mathrm{hand}}_{n,v},
\mathcal{S}^{\mathrm{obj}}_{n}\right)
< \delta_c
\right],
\label{eq:contact_threshold}
\end{equation}
where $\mathbf{x}^{\mathrm{hand}}_{n,v}$ is the $v$-th MANO vertex,
$\mathcal{S}^{\mathrm{obj}}_{n}$ is the object surface, and $\delta_c$ is the
contact threshold.

Although practical, thresholding discards proximity information and makes labels sensitive to $\delta_c$ and geometric accuracy. We therefore retain HACO's binary contact supervision and class-imbalance treatment while predicting distance. Contact can then be obtained either from the contact head or by thresholding the distance prediction. This preserves geometric information when distance labels are available without requiring them for every sample.

\noindent\textbf{Contact loss.}
Following HACO~\cite{jung2025haco}, we use the multi-scale vertex contact
balanced (VCB) loss
\begin{equation}
\begin{aligned}
\mathcal{L}_{\text{contact}}
={}&
\mathcal{L}^{\mathrm{Full}}_{\text{VCB}}
+\mathcal{L}^{\mathrm{High}}_{\text{VCB}}
+\mathcal{L}^{\mathrm{Medium}}_{\text{VCB}}
+\mathcal{L}^{\mathrm{Joint}}_{\text{VCB}} \\
&+0.1\,\mathcal{L}_{\text{Reg}}
+\mathcal{L}_{\text{Smooth}} .
\end{aligned}
\label{eq:contact}
\end{equation}
The VCB terms supervise predictions on the full mesh, two coarsened meshes, and
a joint-level representation. Coarse labels are soft targets computed as
$c_{\text{low}}=V_{\text{reg}}c_{\text{mesh}}$, preserving partial contact
within aggregated regions. $\mathcal{L}_{\text{Reg}}$ limits deviation from the
dataset-level mean contact map, while $\mathcal{L}_{\text{Smooth}}$ encourages
consistency between adjacent vertices. Because all training samples have contact
annotations, no mask is required.

\noindent\textbf{Distance loss.} The distance head mirrors the contact head and predicts the distance from each hand vertex to the object surface. This task is highly imbalanced: only a small fraction of the $778$ MANO vertices are near the contact region, while most are far from the object. We address this imbalance using the three design choices in Eq.~\ref{eq:distpersample}.


\vspace*{-\baselineskip}
\begin{align}
w_{n,i}
&=
\begin{cases}
w_{\mathrm{near}}, & d^{\mathrm{gt}}_{n,i}<\tau_{\mathrm{near}},\\[2pt]
1, & \text{otherwise},
\end{cases} \nonumber\\
\ell_n
&=
\frac{1}{\sum_i w_{n,i}}
\sum_i w_{n,i}
\left|
\log(1+\hat d_{n,i})
-\log(1+d^{\mathrm{gt}}_{n,i})
\right|, \nonumber\\
\mathcal{L}_{\mathrm{dist}}
&=
\frac{\sum_n m_n\,\ell_n}
{\max\!\left(\sum_n m_n,\,1\right)}.
\label{eq:distpersample}
\end{align}

\emph{(i) Near-field reweighting ($w_{n,i}$).}
Vertices with a ground-truth distance below $\tau_{\mathrm{near}}$ are assigned a weight of $w_{\mathrm{near}}=10$, while all other vertices receive unit weight. Without this reweighting, uniformly averaging over all $778$ vertices reduces the contribution of the informative near-contact region by approximately a factor of $15$.

\emph{(ii) Log-space regression ($\ell_n$).}
We regress the predicted and ground-truth distances after applying the transformation $\log(1+d)$. This transformation is well-defined at $d=0$ and maps exact contact to zero. For small distances, $\log(1+d)\approx d$, so variations near the object surface are largely preserved. As distance increases, the slope $1/(1+d)$ decreases, compressing large values and preventing errors at distant vertices from dominating the optimisation. Since the transformation is monotonic, it preserves the ordering of distances rather than discarding far-field information. This retains global distance supervision while placing greater emphasis on the near-contact region.

\emph{(iii) Sample-level masking and mask-normalised averaging ($\mathcal{L}_{\mathrm{dist}}$).}  $\hat d_{n,i}$ and $d^{\mathrm{gt}}_{n,i}$ denote the predicted and ground-truth distances at vertex $i$ of sample $n$, and $m_n\in\{0,1\}$ indicates annotation validity. Each sample loss is normalised by its total vertex weight, making it independent of the number of near-contact vertices. The loss is normalised by the number of annotated samples, so unannotated samples contribute neither loss nor gradients. The denominator is clamped to one for batches without valid annotations.

\noindent\textbf{Pressure loss.} The pressure head predicts a per-vertex pressure
through a sigmoid and is supervised with a masked $L_1$ loss (no near-field
reweighting):
\begin{equation}
\mathcal{L}_{\text{pressure}} \;=\;
   \frac{\sum_n \big|\hat p_n - p^{\text{gt}}_n\big|_{\text{mean}}\,
         m^{\text{press}}_n}{\sum_n m^{\text{press}}_n}.
\label{eq:pressure}
\end{equation}
Ground-truth pressure is normalized by a cap of $20$\,kPa and clipped to
$[0,1]$. Pressure labels come from a single dataset (PressureVisionDB \cite{grady2022pressurevision}),
${\approx}2.88\%$ of the training mixture.

\section{Experiments}
\label{sec:exp}

\subsection{Hand Contact Datasets}

\begin{table}[t]
\centering
\footnotesize
\setlength{\tabcolsep}{2.5pt}
\begin{tabular}{@{}cccc@{}}
\toprule
Dataset & Size & Share & Type \\
\midrule
\multicolumn{4}{@{}l}{\itshape
A. Sensor-measured (high reliability)} \\
PressureVisionDB~\cite{grady2022pressurevision}
& 33,514 & 2.88\% & Pressure \\
\multicolumn{4}{@{}l}{\itshape
B. Multi-view, MoCap (high reliability)} \\
TACO~\cite{liu2024taco}
& 153,097 & 13.17\% & Manipulation \\
OakInk2~\cite{zhan2024oakink2}
& 147,553 & 12.69\% & Manipulation \\
ARCTIC~\cite{fan2023arctic}
& 147,238 & 12.66\% & Manipulation \\
Decaf~\cite{shimada2023decaf}
& 65,322 & 5.62\% & Hand--face \\
MANUS~\cite{pokhariya2024manus}
& 20,117 & 1.73\% & Grasping \\
RICH~\cite{huang2022rich}
& 15,854 & 1.36\% & Human--scene \\
Hi4D~\cite{yin2023hi4d}
& 2,889 & 0.25\% & Human--human \\
\multicolumn{4}{@{}l}{\itshape
C. RGB-D + fitting (medium)} \\
OakInk~\cite{yang2022oakink}
& 129,659 & 11.15\% & Grasping \\
HOI4D~\cite{liu2022hoi4d}
& 84,428 & 7.26\% & Manipulation \\
HO3D~\cite{hampali2020honnotate}
& 62,779 & 5.40\% & Manipulation \\
HOCap~\cite{wang2025hocap}
& 62,075 & 5.34\% & Manipulation \\
H2O3D~\cite{hampali2022keypoint}
& 59,870 & 5.15\% & Manipulation \\
H2O~\cite{kwon2021h2o}
& 33,467 & 2.88\% & Manipulation \\
DexYCB~\cite{chao2021dexycb}
& 29,656 & 2.55\% & Grasping \\
HOGraspNet~\cite{cho2024hograspnet}
& 29,243 & 2.51\% & Grasping \\
\multicolumn{4}{@{}l}{\itshape
D. Synthetic / monocular reconstruction} \\
ObMan~\cite{hasson2019obman}
& 85,688 & 7.37\% & Grasping \\
MOW~\cite{cao2021mow}
& 350 & 0.03\% & Reconstruction \\
\midrule
Total (training pool) & 1,162,799 & 100\% & --- \\
\midrule
\multicolumn{4}{@{}l}{\bfseries Excluded datasets} \\
\cmidrule(lr){1-4}
Dataset & Size & \multicolumn{2}{l}{Exclusion reason} \\
\midrule
GigaHands~\cite{fu2025gigahands}
& 665,176
& \multicolumn{2}{l}{unreliable, mostly no contact} \\
EgoTouch~\cite{zhou2026touchanything}
& 127,676
& \multicolumn{2}{l}{gloved hands, not bare skin} \\
InterHand2.6M~\cite{moon2020interhand}
& 37,170
& \multicolumn{2}{l}{hand--hand contact only} \\
GRAB~\cite{taheri2020grab}
& 29,612
& \multicolumn{2}{l}{mocap only, no real images} \\
HOT3D~\cite{banerjee2025hot3d}
& 13,456
& \multicolumn{2}{l}{crop/side-matching errors} \\
Open3DHOI~\cite{wen2025open3dhoi}
& 2,561
& \multicolumn{2}{l}{no contact after SMPL-X fitting} \\
EgoPressure~\cite{zhao2025egopressure}
& 2,569
& \multicolumn{2}{l}{contact--image misalignment} \\
HIC~\cite{tzionas2016hic}
& 228
& \multicolumn{2}{l}{hand--hand; 39 test samples} \\
PROX~\cite{hassan2019prox}
& 178
& \multicolumn{2}{l}{SMPL-X mapping unreliable} \\
OpenTouch~\cite{song2025opentouch}
& 146
& \multicolumn{2}{l}{unresolvable crop errors} \\
\midrule
\multicolumn{4}{@{}l}{\bfseries Held-out evaluation only (never sampled in training)} \\
\cmidrule(lr){1-4}
ContactPose~\cite{brahmbhatt2020contactpose}
& 2,192 & --- & Grasping, test only \\
EPIC-Contact~\cite{bansal2026hopformer}
& 237 & --- & Manipulation, test only \\
\bottomrule
\end{tabular}
\caption{Dataset audit details 18 retained datasets, including their share of the training pool and primary interaction types. It also lists excluded datasets with reasons for their omission. ContactPose and EPIC-Contact (bottom block) are held out for evaluation.}
\label{tab:dataset-audit}
\end{table}

\label{sec:data}
The estimation of dense hand contact is an emerging area of research, with this study offering a significant contribution through a comprehensive review of publicly available hand-contact datasets up to July 2026. The framework assesses 30 datasets, encompassing around 4 million samples, and amalgamates 18 that meet stringent quality-control standards into a unified training corpus containing 1.16 million samples. The ten excluded datasets failed due to various issues, such as inconsistent or misaligned contact annotations (notably GigaHands, EgoPressure, and HOT3D), domain disparity with bare-hand–object interactions (EgoTouch, InterHand2.6M, and HIC), and absence of real-world imagery (GRAB). To promote reproducibility, the detailed audit is published to assist future researchers in avoiding the burdensome process of dataset screening.

The study identifies three primary sources of contact labels that shape the design of the corpus. The first source involves geometric determination through thresholding the distance between hand vertices and the object surface, as seen in datasets like ObMan \cite{hasson2019obman} and DexYCB \cite{chao2021dexycb}, which were initially created for hand–object reconstruction. The second method includes direct measurements or annotations, such as thermal imprinting in ContactPose and manual labeling, along with pseudo-ground-truths derived via multiple view optimization in Decaf \cite{shimada2023decaf}. The third source consists of automated propagation of human-annotated contact labels through pipelines, exemplified by MOW \cite{cao2021mow} and EPIC \cite{bansal2026hopformer} Contact. Historically, contact estimation has been tied to reconstruction tasks, leading to the prevalence of geometrically derived labels in existing datasets and the current corpus.

\subsection{Evaluation}

\subsubsection{Unified threshold measurement}
To assess progress while limiting MOW's format bias, we use the two datasets with the most accurate annotations as our primary test sets: EPIC-Contact \cite{bansal2026hopformer}, the latest dataset for contact estimation, and ContactPose \cite{brahmbhatt2020contactpose}, which provides accurate contact labels from thermal imprints. Neither HACO nor DenseHap was trained on these datasets, enabling a fair zero-shot comparison. For in-distribution evaluation, we retain MOW from HACO's original evaluation framework and add ARCTIC \cite{fan2023arctic}, which provides relatively accurate contact annotations. We exclude HIC \cite{tzionas2016hic} and Hi4D \cite{yin2023hi4d} from main evaluation because they focus primarily on hand-to-hand and hand-to-body interactions, which fall outside our evaluation scope.

\begin{table}[h]
    \centering
    \footnotesize
    \setlength{\tabcolsep}{3pt}
    \renewcommand{\arraystretch}{1.08}

    \begin{tabular}{lccccccccc}
        \toprule
        & & \multicolumn{4}{c}{DenseHap}
            & \multicolumn{4}{c}{HACO} \\
        \cmidrule(lr){3-6}
        \cmidrule(lr){7-10}
        Dataset & $N$
            & P & R & F1 & F1$^{\star}$
            & P & R & F1 & F1$^{\star}$ \\
        \midrule
        EPIC-Contact
            & 2272
            & .453 & .622 & \textbf{.511}
            & \textbf{.525}
            & .386 & .595 & .428 & .459 \\

        ContactPose
            & 2192
            & .606 & .812 & \textbf{.673}
            & \textbf{.675}
            & .565 & .680 & .588 & .599 \\

        MOW
            & 92
            & .432 & .926 & \textbf{.555}
            & \textbf{.570}
            & .525 & .607 & .522 & .567 \\

        ARCTIC
            & 3000
            & .675 & .792 & \textbf{.703}
            & \textbf{.713}
            & .516 & .787 & .682 & .684 \\
        \midrule
        Mean
            & --
            & .542 & .788 & \textbf{.600} & \textbf{.621}
            & .498 & .667 & .555 & .578 \\
        \bottomrule
    \end{tabular}

    \caption{
        Dense contact estimation under the Skip protocol.
        We report per-vertex precision (P), recall (R), and F1 score;
        F1 denotes performance at the default threshold ( 0.52), while
        F1$^{\star}$ denotes performance after readout and operating-point
        selection.
    }
    \label{tab:main}
\end{table}

\label{sec:main}
Table~\ref{tab:main} summarizes the results. DenseHap outperforms HACO on all four datasets, with the largest gains on the most accurately annotated benchmarks. Under zero-shot evaluation, F1 improves by 8.5 points (+14.5\% relative) on thermally measured ContactPose and by 8.3 points (+19.4\%) on EPIC-Contact. The in-distribution gains on MOW (+3.3 points) and ARCTIC (+2.1 points) are smaller but consistent. The stronger performance on unseen benchmarks indicates estimation quality rather than overfitting to a particular label distribution.

Furthermore, DenseHap(distance head) enhances both aspects of predictive efficacy: mean recall escalates from $.667$ to $.788$, while mean precision advances from $.498$ to $.542$. The largest precision gain occurs on ARCTIC, from $.516$ to $.675$. MOW illustrates the caveat discussed earlier: DenseHap achieves near-complete recall ($.926$) but low precision ($.432$), because matching MOW's exaggerated labels requires predicting contact well beyond the true touch region. This setting favors broad predictions and penalizes precise ones. We therefore treat MOW as a historical benchmark rather than a primary measure of contact quality.
\subsubsection{Broader Evaluation Using Optimal Thresholds}

Various techniques utilise distinct thresholds to convert their predictions into binary outcomes for each dataset. To facilitate an equitable comparison, we vary the decision threshold between 0.3 and 0.8 and present the optimal performance attained by each technique on the identical problem. The comprehensive findings are described below.

\begin{table}[h]
\centering
\footnotesize
\setlength{\tabcolsep}{1.5pt}
\begin{tabular}{@{}lcccccc@{}}
\toprule
Method & MOW & ContactPose & EPIC & ARCTIC & RICH & Hi4D \\
\midrule
POSA-hand ~\cite{hassan2021posa} $\dagger$
& .408 & -- & -- & -- & .398 & .315 \\
ContactPrompt~\cite{jung2026contactprompt} $\dagger$
& .526 & -- & -- & -- & -- & -- \\
DECO~\cite{tripathi2023deco}
& .395\rlap{$^{\dagger}$} & .513 & .492 & .365
& .454 & .304\rlap{$^{\dagger}$} \\
BSTRO~\cite{huang2022bstro}
& .380\rlap{$^{\dagger}$} & .155 & .382 & .290
& .645 & .420\rlap{$^{\dagger}$} \\
HACO~\cite{jung2025haco}
& .522 & .599 & .459 & .684 & .781 & \textbf{.474} \\
\midrule
JointHand (ours)
& \textbf{.571}\rlap{\textsuperscript{\scriptsize d}}
& \textbf{.675}\rlap{\textsuperscript{\scriptsize d}}
& \textbf{.524}\rlap{\textsuperscript{\scriptsize c}}
& \textbf{.713}\rlap{\textsuperscript{\scriptsize d}}
& \textbf{.799}\rlap{\textsuperscript{\scriptsize c}}
& .424\rlap{\textsuperscript{\scriptsize c}} \\
\bottomrule
\end{tabular}
\caption{\textbf{Dense hand contact estimation, per-vertex F1 at each
method's own best threshold (Skip protocol).}
Our result uses the distance head (d) per dataset; EPIC, RICH, and Hi4D
use only the contact head (c) because they lack distance ground truth.
$\dagger$ denotes HACO-reported results. Other results were evaluated
locally under the same protocol.}
\label{tab:contact_best}
\end{table}

Even under this ideal threshold assessment framework, our technique consistently surpasses other methods across every dataset except Hi4D. In the in-distribution datasets, our approach realises an average enhancement of 3–5\% compared to HACO, the current leading model. In the out-of-distribution datasets, the mean enhancement approaches 10\%. Considering HACO's robust alignment with these datasets, these findings suggest that our model has enhanced generalisation to unfamiliar distributions.


\begin{figure}[h]
\begin{center}
 \includegraphics[width=1\linewidth]{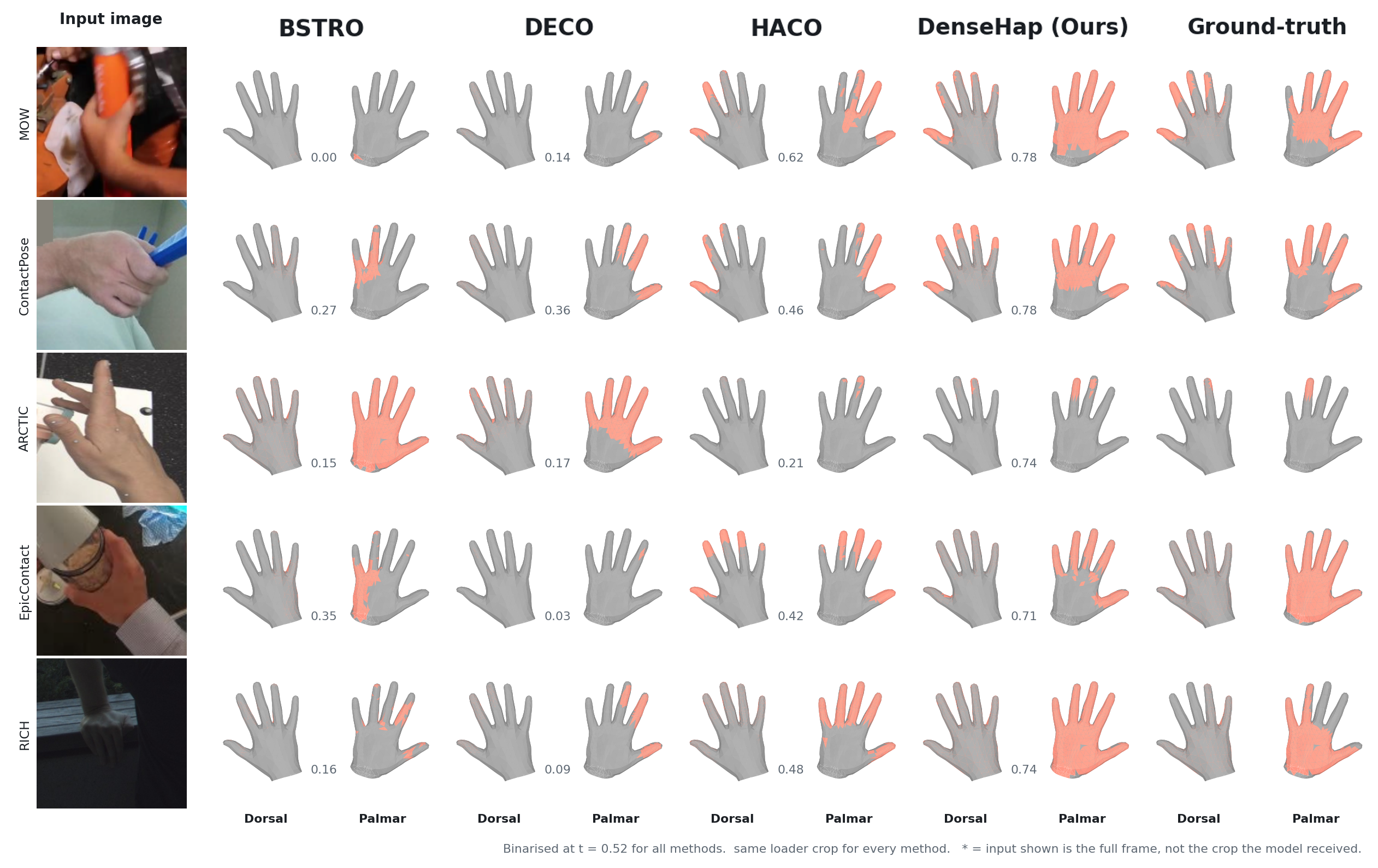}
 \vspace{-5pt} 
\end{center}
 \vspace{-5pt} 
   \caption{Qualitative comparison between the proposed and the baseline method across five datasets.}
   
\label{fig:show}
\end{figure}

Figure \ref{fig:show} illustrates examples generated by several models. The contact estimation for models such as BSTRO and DECO appears to be relatively imprecise. HACO enhances performance, although it frequently applies a rudimentary colouration to both sides of the hand without differentiating between the palm and the dorsal side. Furthermore, it demonstrates some interference and faces challenges in accurately detecting specific outlier instances. While DenseHap does well on clear photos, it struggles with overly dark or ambiguous ones.

\subsubsection{Distance and Force measurement}

Alongside our main goal, we also performed comparison analyses on the distance and force heads. The findings are delineated below.
\begin{table}[h]
\centering
\small
\setlength{\tabcolsep}{7pt}
\begin{tabular}{@{}lcccc@{}}
\toprule
& \multicolumn{4}{c}{H$\rightarrow$O Distance MAE (mm) $\downarrow$} \\
\cmidrule(lr){2-5}
Method & All & $<10$ & $10$--$20$ & $\geq 20$ \\
\midrule
InterField-SF ~\cite{fan2023arctic}     & \textbf{9.47} & 7.92          & 7.62          & \textbf{10.51} \\
Our distance head & 11.49         & \textbf{3.70} & \textbf{4.12} & 16.21 \\
\bottomrule
\end{tabular}
\caption{Continuous hand-to-object distance on ARCTIC.
MAE is computed over 778 MANO vertices with distances capped at 100\,mm.}
\label{tab:distance}
\end{table}

Table \ref{tab:distance} primarily contrasts our model with InterField utilising the ARCTIC dataset. Particularly, our distance head features comprehensive near-field optimisation, providing a distinct benefit in assessing proximate contacts at ranges under 20 mm. InterField exhibits superior performance for distances above 20 mm, hence yielding optimal results among all outcomes. Nevertheless, the majority of hand-contact datasets establish 10 mm as the principal criterion for defining contact, with a few exceptions that set the threshold higher than 10 mm. Consequently, precise distance measurement within 20 mm is crucial for high-accuracy hand-contact datasets. Distances over 20 mm typically do not influence touch recognition. These findings convincingly corroborate the architecture presented in Equation \ref{eq:distpersample} as well as the functionality of the distance head.

There is currently no efficient MANO-based method for predicting pressure on a per-vertex
basis. The closest work, WristP2~\cite{xi2026wristp2}, is based on wrist-mounted imagery
and so differs substantially from our setting. We therefore use PressureVision as our main
reference point; the results are shown in Table~\ref{tab:pressure}.

\begin{table}[h]
\centering
\footnotesize
\setlength{\tabcolsep}{5pt}
\begin{tabular}{@{}lccc@{}}
\toprule
Method & VolIoU\,$\uparrow$ & MAE\,(Pa)\,$\downarrow$ & C-IoU\,$\uparrow$ \\
\midrule
\multicolumn{4}{@{}l}{\textit{Image space, $480\times384$\,px}} \\
PressureVision & 0.413 & 4246 & 0.558 \\
\midrule
\multicolumn{4}{@{}l}{\textit{Mesh space, MANO-778}} \\
Failed blank control, $\ell_1$ loss$^{\dagger}$
    & 0.000 & 6306 & 0.000 \\
Ours (full model), BCE\,+\,proxy
    & \textbf{0.491} & \textbf{2840} & \textbf{0.597} \\
\bottomrule
\end{tabular}
\caption{\textbf{Hand pressure prediction on PressureVisionDB}, using the same
six test identities. Cross block results are reference due to different
output spaces. Metrics use the support
$\{\hat p>1\,\text{kPa}\}\cup\{p>1\,\text{kPa}\}$.
$^{\dagger}$Failed blank control producing all-zero pressure.
\textbf{Caveat:} all checkpoints saw these identities during training.}
\label{tab:pressure}
\end{table}

Given PressureVision's different formulation, which predicts pressure through
nine discrete categories in image space, the resulting measures should be
viewed as a reference rather than a directly equivalent evaluation. For each
method, we report the mean absolute error in its native output space over the
union of predicted and ground-truth contact support. The all-zero masked
$\ell_1$ run is included only as a failed blank control, since it collapsed
during training. Our actual BCE\,+\,proxy model avoids this trivial solution,
indicating that the pressure-head objective plays a significant role in learning
and achieves performance comparable to PressureVision under this adapted
protocol.

\section{Ablation Study}

\noindent\textbf{Model design Comparison.}
\begin{table}[t]
\centering
\footnotesize
\setlength{\tabcolsep}{3pt}
\begin{tabular}{@{}l ccccc@{}}
\toprule
Contact head & MOW & HIC & RICH & Hi4D & Mean \\
\midrule
\multicolumn{6}{@{}l}{\textit{A. Each vertex sees only its own image patch}}\\
MLP, 1 hidden layer          & .462 & .109 & .527 & .516 & .404 \\
MLP, 2 hidden layers         & .464 & .107 & .527 & .515 & .403 \\
MLP, 5 hidden layers         & \bf.502 & \bf.168 & \bf.551 & \bf.519 & \bf.435 \\
Graph network on the mesh    & .477 & .153 & .536 & .517 & .421 \\
\midrule
\multicolumn{6}{@{}l}{\textit{B. All vertices share one summary of the image}}\\
Average the image features   & .511 & .183 & .700 & \bf.523 & .479 \\
Learn one summary token      & .504 & .191 & .787 & \bf.524 & .502 \\
\quad $+$ focal loss         & \bf.522 & \bf.271 & \bf.792 & .521 & \bf.526 \\
\midrule
\multicolumn{6}{@{}l}{\textit{C. Each vertex looks up the image itself (attention)}}\\
One shared query             & .522 & .263 & .781 & .565 & .533 \\
\quad $+$ more training data & .529 & \bf.332 & \bf.799 & .524 & \bf.546 \\
\quad $+$ rebalanced sampling & \bf.535 & .286 & .752 & \bf.613 & \bf.546 \\
\quad $+$ fine-tune backbone & .501 & .274 & .785 & .566 & .531 \\
\addlinespace[2pt]
One query per vertex         & .519 & .230 & .756 & -- & -- \\
\quad $+$ also predict distance & .460 & .177 & .719 & .322 & .419 \\
\quad $+$ also predict pressure & .483 & .174 & .757 & .324 & .435 \\
Queries learned from scratch & .495 & .330 & .723 & .605 & .538 \\
\midrule
\multicolumn{6}{@{}l}{\textit{D. Vertices inherit their hand part's prediction}}\\
Predict 84 hand parts        & \bf.510 & \bf.246 & .724 & \bf.413 & \bf.473 \\
\quad $+$ also predict distance & .489 & .188 & \bf.749 & .370 & .449 \\
\bottomrule
\end{tabular}
\caption{\textbf{Contact-head architectures}, grouped by what each vertex's prediction is
conditioned on. MLP hidden layer widths are $64$, $256$ and $128$; ``more training data'' is
$14$ datasets instead of $13$; ``rebalanced sampling'' draws each dataset in proportion to
$\sqrt{N}$; ``queries learned from scratch'' replaces the
MANO-seeded queries with DINOv3. Single threshold $t{=}0.52$ for all
datasets, per-sample F1, non-contacting samples scored as $0$; bold marks the best in each
group.}
\label{tab:arch}
\end{table}
In Table \ref{tab:arch} we compare contact-head architectures according to how they decode the 778 vertex-level logits. All models are evaluated using the same protocol as HACO, with a shared threshold of \(t=0.52\) across all datasets and \(\textsc{None}\rightarrow 0\) aggregation.

The results favor architectures that aggregate global information over independent local prediction. Projecting each hand vertex onto backbone features and predicting contact from a local patch is insufficient because contact states are spatially correlated across vertices and local evidence is ambiguous under occlusion. Global cues such as object geometry, scale, and interaction type are needed to distinguish narrow contacts, such as grasping a pencil, from broad planar contacts, such as touching a table. Cross-attention addresses these limitations by retrieving vertex-specific evidence from a global representation.


The best per-vertex model achieves a mean F1 score of 0.435, while global-vector decoding and learned queries yield scores of 0.546 and 0.538, respectively. The one-token cross-attention variant scores 0.533, surpassing several methods including max pooling and CLS-based self-attention. Performance is influenced by dataset composition, sampling, and augmentation, with up to five F1 point variations within the same model family. The 14-dataset model excels on HIC and RICH, while corrected square-root sampling with augmentation performs best on MOW and Hi4D. Effective contact estimation requires global decoding and adaptable training strategies to manage variations in interactions and dataset bias.

\noindent\textbf{Contact Head.}
Although scaling laws suggest that larger models and datasets improve performance, our decoupled framework depends primarily on encoder capacity and embedding quality. Since larger contact heads may overfit the relatively small HACO dataset, we ablate head dimensions while keeping the cross attention architecture and all other settings fixed.
\begin{table}[h]
\centering\footnotesize\setlength{\tabcolsep}{5pt}
\begin{tabular}{@{}lrcccc@{}}
\toprule
Contact head & Params & MOW & HIC & RICH & Hi4D \\
\midrule
\multicolumn{6}{@{}l}{\emph{(a) Cross-attention, Skip protocol, best $t$}}\\
Full size    & 25\,M & .547 & \bf.348 & \bf.799 & .699 \\
Small        & 1\,M  & \bf.550 & .266 & .778 & .698 \\
Tiny         & 60\,k & .533 & .265 & .781 & \bf.701 \\
\cmidrule(l{2pt}){2-6}
\emph{spread} & $400\times$ & .017 & \bf.083 & .021 & .003 \\
\midrule
\multicolumn{6}{@{}l}{\emph{(b) Per-vertex MLP, non-contact scored $0$, $t{=}0.52$}}\\
1 hidden layer  & 0.07\,M & .462 & .109 & .527 & .516 \\
2 hidden layers & 0.35\,M & .464 & .107 & .527 & .515 \\
5 hidden layers & 5.1\,M  & \bf.502 & \bf.168 & \bf.551 & \bf.519 \\
\cmidrule(l{2pt}){2-6}
\emph{spread} & $70\times$ & .040 & .061 & .024 & .004 \\
\bottomrule
\end{tabular}
\caption{\textbf{Head capacity, architecture held fixed.} Cross-attention widths are
$1024/256/64$ with $6/2/1$ layers; MLP hidden widths are $64$, $256$ and
$128$. \emph{spread} is the gap between the best and worst row. Protocol is
constant within a panel but differs between them, so the two panels are not comparable
cell by cell.}
\label{tab:capacity}
\end{table}
As shown in Table \ref{tab:capacity}, cross-attention performance remains stable on MOW, RICH, and Hi4D even when head parameters are reduced by 400×, indicating that head capacity is not critical. On HIC, performance decreases as capacity grows, suggesting overfitting. In contrast, MLP performance varies substantially with model size, while the largest model still perform worse than cross attention head. Overall, cross-attention performs better, while smaller heads can improve generalisation on specific interaction patterns.

\noindent\textbf{Component ablations.}
\label{sec:ablation}
Table~\ref{tab:ablation} isolates the contribution of each component on the
two cleanest benchmarks, ContactPose (zero-shot) and ARCTIC. The baseline
variant already couples 2D-projected vertex queries with distance
supervision and surpasses the HACO re-run on both datasets, confirming the
central claim of our loss design: supervising the continuous distance
underlying the binary labels is itself a stronger training signal than
binary contact alone. Conditioning the contact prediction on the estimated
distance adds a further $+.011$ on ContactPose. Dataset-balanced BCS
weighting in isolation is slightly unfavorable on these clean-label sets
($-.005$ on ContactPose), which is expected: BCS shifts capacity toward
rare, noisier datasets and pays off in breadth rather than on any single
benchmark. The full model, which uses three parameter-disjoint trunks with detached proximity injection, recovers this cost and reaches the best result on both datasets (.673 and .703). This shows that gradient isolation lets contact, distance, and pressure specialize without interfering, while the injected proximity cues still reach the contact head at inference.

\begin{table}[h]
\centering
\footnotesize
\begin{tabular}{@{}lcc@{}}
\toprule
Variant & ContactPose & ARCTIC \\
\midrule
HACO \cite{jung2025haco} & .588 & .682 \\
\midrule

2D proj.\ + distance supervision & .653 & .687 \\
+ distance-conditioned contact & .664 & .688 \\
+ BCS weighting & .659 & .686 \\
Full: 3 trunks + proximity injection & \textbf{.673} & \textbf{.713} \\
\bottomrule
\end{tabular}

\caption{Ablation study, per-vertex F1 via distance binarization on
ContactPose (zero-shot) and ARCTIC. Each row adds one component to the
previous. All variants surpass HACO; the full model performs best on both
datasets.}
\label{tab:ablation}
\end{table}

\subsection{Runtime Performance}
\label{sec:speed}

\begin{table}[h]
\centering
\footnotesize
\setlength{\tabcolsep}{5pt}
\begin{tabular}{@{}lccc@{}}
\toprule
Method & Params & FPS & Latency (ms) \\
\midrule
ViT-H backbone only      & ${\approx}600$M  & 43.5 & $22.99 \pm 0.28$ \\
WiLoR (recon.\ only)     & ${\approx}640$M  & 31.8 & $31.49 \pm 0.35$ \\
HACO (contact only)      & ${\approx}630$M  & 30.0 & $33.32 \pm 1.11$ \\
WiLoR\,$+$\,HACO cascade & ${\approx}1.2$B  & 15.4 & $64.81 \pm 3.25$ \\
DenseHap (ours)         & ${\approx}672$M  & 26.1 & $38.37 \pm 4.93$ \\
\bottomrule
\end{tabular}
\\
\caption{Efficiency on a single GH200 (bf16, batch size 1, mean over 200
runs). Sharing one backbone pass makes DenseHap ${\approx}1.7\times$
faster than the WiLoR\,$+$\,HACO cascade; its ${\approx}40$M trainable
head is ${\sim}15\times$ smaller than the backbone and adds only
$6.9$\,ms over reconstruction-only WiLoR.}
\label{tab:speed}
\end{table}


We also evaluate model efficiency. Running WiLoR and HACO independently introduces redundant computation, as both rely on large backbones to encode the same image (Table~\ref{tab:speed}). DenseHap avoids this redundancy by sharing a single visual representation across reconstruction and interaction reasoning, providing a better balance between throughput and functionality than the cascaded system.

Most of DenseHap’s overhead comes from a lightweight task-specific head rather than a second image encoder. It retains efficient reconstruction-only inference while additionally predicting contact, distance, and pressure. Consequently, DenseHap is 1.7× faster than the WiLoR, HACO cascade, making it suited for large-scale video annotation.


\section{Conclusion and Limitations}
In conclusion, we have developed DenseHap, a comprehensive framework that, in a single forward pass, provides a detailed per-vertex contact, distance, and 3D hand mesh from a single RGB image. After constructing a clean dataset from 28 relevant datasets, we integrated three parameter-independent vertex-query cross-attention trunks into a static WiLoR backbone and trained them using a multiple loss function design to address class imbalance. Our model achieves a throughput of approximately 1.7 $\times$ that of a reconstruction-contact cascade and improves the F1 score by approximately 10\% compared to the current state of the art across zero-shot datasets. 

\noindent\textbf{Limitations.} Despite the performance improvements, our method still suffers from some limitations. Firstly, the model performance can be highly affected under the presence of severe occlusions. In addition, given that our main objective is to predict hand object contact and pressure, our model is not tailored for hand-hand and hand-body interactions. Nevertheless, both limitations could be easily addressed with the use of additional datasets that include heavier occlusions and hand-body interactions.  

{
    \small
    \bibliographystyle{ieeenat_fullname}
    \bibliography{main}
}

\end{document}